\documentclass[10pt,letterpaper]{article}

\usepackage{cogsci}

\cogscifinalcopy 
\usepackage{graphicx}
\usepackage{pslatex}
\usepackage{apacite}
\usepackage{float} 

\title{Understanding the Countably Infinite: Neural Network Models of the Successor Function and its Acquisition}

\author{{\large \bf Vima Gupta, Sashank Varma} \\({vima.gupta, varma)}@gatech.edu \\
Georgia Institute of Technology}

\begin{document}

\maketitle

\begin{abstract}
As children enter elementary school, their understanding of the ordinal structure of numbers transitions from a memorized count list of the first 50-100 numbers to knowing the successor function and understanding the countably infinite. We investigate this developmental change in two neural network models that learn the successor function on the pairs $(N, N+1)$ for $N \in (0, 98)$. The first uses a one-hot encoding of the input and output values and corresponds to children memorizing a count list, while the second model uses a place-value encoding and corresponds to children learning the language rules for naming numbers. The place-value model showed a predicted drop in representational similarity across tens boundaries. Analysis of the latent representation shows that counting across a tens boundary can be understood as a vector operation in 2D space, where the numbers with the same tens place are organized in a linearly separable manner, whereas those with the same ones place are grouped together. A curriculum learning simulation shows that, in the expanding numerical environment of the developing child, representations of smaller numbers continue to be sharpened even as larger numbers begin to be learned. These models set the stage for future work using recurrent architectures to move beyond learning the successor function to simulating the counting process more generally, and point towards a deeper understanding of what it means to understand the countably infinite.

\end{abstract}

\section{Introduction}

The foundations of mathematics rest in part on the Peano axioms that define the natural numbers \citeA{giaquinto2001knowing}. A key element of these foundations is the successor function $S(N)$, which given the natural number $N$ as input gives the natural number $N+1$ that immediately follows it. It is a challenge for Artificial Intelligence (AI), machine learning (ML), and cognitive science to explain how intelligent systems learn the successor function and, more generally, come to understand the countable infinity of the natural numbers.

Cognitive and developmental scientists have investigated how children learn about the natural numbers \citeA{carey2019ontogenetic, doi:10.1080/09515080802285354, SPELKE2017}. To do so, they must master multiple related concepts. They must learn to count, which entails constructing a ``count list'' that orders the numbers they know up to a maximum; this maximum increases over development. The question of how this list is bootstrapped -- how the natural numbers 1-3 are learned -- is a matter of debate \citeA{doi:10.1080/09515080802285354, SPELKE2017}.

By whatever process the initial natural numbers are bootstrapped, children extend them as they construct their ``count list'' -- again, an ordered sequence with a maximum that increases over development \citeA{carey2019ontogenetic}. An important conceptual breakthrough occurs between the ages of 3.5 and 4, when children learn to \emph{precisely} count the number of items in a set (within their counting range) or generate sets of a given number (again, within their counting range). This links the ordinal and cardinal senses of number, which is important because knowing how counting related to numbers marks children as cardinal principle (CP) knowers \citeA{carey2019ontogenetic, Sarnecka2013, Spaepen2018}.

Another concept that children must master is that of the successor function $S(N)$.
\citeA{Cheung2017ToIA} asked CP-knowers ages 4-7 (1) to generate the successor of numbers in their count list and (2) justify whether all numbers have successors. Only children ages 5.5 to 6 years could correctly answer these questions, and thus demonstrate a proper understanding of the successor function and, by extension, of the countable infinity of the natural numbers.

There is disagreement on the order in which children develop these concepts. Some researchers argue that understanding of the successor function comes before, and supports, becoming a CP knower \citeA{carey2019ontogenetic}. Others propose that children become CP knowers first \citeA{Davidson2012, Spaepen2018}, with understanding of the successor function lagging behind by 2 years \citeA{Cheung2017ToIA}.

Here, we focus on the successor function and its acquisition. Multiple researchers have proposed that learning to count drives induction of the successor function \citeA{Cheung2017ToIA, carey2019ontogenetic, SARNECKA2008662}. And in turn, language guides the construction of the count list~\citeA{Hurford2011}. Learning the successor function depends on learning the rules for naming numbers in one's native language \citeA{Guerrero2020IsTT}. For most languages, these rules reflect the structure of the place-value symbol system \citeA{Pollmann1996TheLU}, though languages differ in the transparency of this correspondence \citeA{Miller1995PreschoolOO}. Number naming rules, like those of the place-value symbol system, are inherently recursive \citeA{Schneider2020DoCU}. This structure bootstraps learning of the successor function. Importantly, the successor function must be sensitive to the ``boundaries'' of place-value notation. For example, computing $S(29)$ may be more difficult for children to learn than computing $S(28)$.

A key outstanding question is by what mechanisms do children discover the infinite structure of the natural numbers~\citeA{Piantadosi2023}? Answering this question requires the development of formal theories and models. \citeA{Piantadosi2023} took a ``language of thought'' approach to addressing this research goal. Their Bayesian model learns programs for mapping sets (e.g., ``o o o'') to numbers (e.g., ``3''). The programs are compositions of primitive logical, set-theoretic, sequential, and recursive functions. Although the space of possible programs is huge, in practice, their model is able reproduce some of the findings in nth counting development of CP-knowers. A follow-up model,~\citeA{Piantadosi2023}, incorporated an approximate sense of number into the otherwise crisp symbolic program representations of the earlier model. 

In this paper, we use a neural network approach to address the question of how children learn the successor function. For more than a decade, researchers have been evaluating the cognitive and developmental alignment of ML models to the data on number development. Most of this work has involved vision models and the perception and representation of the numerosities of sets of objects. \citeA{Stoianov2012EmergenceOA} trained a multi-later perceptron on artificially generated images of varying numerosity and observed units with psychophysical response functions like those observed in monkey parietal cortex; see also \citeA{Zorzi2018}. \citeA{Testolin2020} generalized this approach and examined the question of number development. They found a sharpening of number representations over training similar to that observed in children over development. Subsequent research has investigated the  latent number representations of convolutional neural network and transformer models trained on ImageNet \citeA{Boccato2021, Kim2021, Nasr2021}, finding hidden layer units tuned to specific numerosities. More recently, researchers have begun probing the number representations of Large Language Models (LLMs). For example, \citeA{Shah2023} investigated the representations of numbers expressed as digits or words in transformer-based models, finding the behavioral signatures indicative in humans of a log-compressed mental number line. Our work differs from these efforts in focusing on the development of counting and the successor function.

In particular, we are not asking an engineering question. In a technical sense, it is a trivial problem for a neural network to learn to compute the successor function for numbers in a fixed range. Current research is using LLMs to solve much more difficult mathematical problems \citeA{Drori_2022, Zhang2023AIFM}. Rather, we are asking a scientific question: How do children learn the successor function, what latent number representations make this possible, and, more generally, what does it mean to come to understand the countably infinite?

In this paper, we investigate two types of numerical encodings, namely one-hot encoding through a \emph{count-list model} and two-hot encoding through a \emph{place-value model}. These models are lightweight multi-layer perceptrons where the number of nodes in the input/output layers scale linearly with the domain and range of the successor function to be learned. Psychologically, the count-list model might correspond to a younger child who has memorized the count list but who lacks a generative understanding of number-naming rules and their inherent place-value structure; this latter, more generative understanding is captured by the place-value model. The motivation is to investigate whether these two models can understand and generalize the successor function. In addition to the models' overall accuracy, we analyze the similarity between the embeddings learned for each number and its successor. Because learning of place-value boundaries is critical to naming the successor of a number, we conduct a further analysis of the embeddings learned across place-value boundaries. 

To preview the results, the place-value model shows the predicted drop in representational similarity across tens boundaries and further reveals three important properties.  First, counting across a tens boundary can be understood as a vector operation in 2D space.
Also in this 2D space, numbers with the same tens place cluster in a linearly separable manner, whereas those with the same ones place exhibit grouping.
Finally, a \emph{curriculum learning} simulation shows that, in the expanding numerical environment of the developing child, the representations of smaller numbers continue to be sharpened even as larger numbers begin to be learned.

\section{Methodology}
\label{sec:methods}
The current study explored two models of the successor function, the count list model and the place-value model, on a toy data set that maps $D:[0,98] \rightarrow R:[1,99]$. We chose this range because in the \citeA{Cheung2017ToIA} study, for the highest counting children, the average maximum value in their count lists was 99.5. This indicates that knowing the natural numbers $1-99$ is generally sufficient for learning the successor function. The count list model investigated whether a one-hot encoding representation of number is sufficient for learning the successor function. It is a multi-layer perceptron with 99 input units, one hidden layer of 8 units, and 99 output units with a softmax activation on the final layer. Learning the successor function in this model corresponds to single-label multi-class classification. This model was trained for 2500 epochs with a batch size of 1 using the KL divergence loss function. 
The final model architectures were determined after extensive experimentation with the number and size of the hidden layers. We also performed hyper-parameter tuning for parameters like the learning rate, activation function, and number of training epochs. Through experimentation with various loss functions such as Wasserstein distance and negative log likelihood loss, activation functions like SiLU, ReLU and Softmax, and having different number of nodes in the hidden layer, we picked the best model architecture based on loss convergence trends and overall model accuracy.

The place-value model investigated learning of a more sophisticated representation of number, one with a place-value encoding. The multi-layer perceptron has 20 input units, three hidden layers of 8 units each, and 20 output units. It uses a ReLU activation in the intermediate layers and a sigmoid activation in the final layer. Once again, we conducted an extensive grid search by varying the number of hidden layers and activation functions (softmax, ReLU and sigmoid) and learning rate to arrive upon the final model architecture and training strategy.

The input and output layers each include 10 units that one-hot encode the tens place and 10 that one-hot encode the ones place. Unlike the count list model, learning the successor function in the place-value model corresponds to multi-label, multi-class classification; hence the different architecture. The place-value model was trained for 5000 epochs with a batch size of 1 using the cross-entropy loss function.

We additionally took a curriculum learning approach to more closely simulate numerical development. These simulations, which exclusively used the place-value model, are described in the later section.

We ran 25 simulations of each model by varying the random seed and statistically analyzed mean performance. For each simulation, we randomly sampled 80 percent of the overall data set to form the training data; the remaining 20 percent comprised testing data. Our experiments used the standard AI/ML approach to model verification in that they included separate training and testing (but no validation) sets. We also constructed loss convergence plots to determine the number of epochs for training to avoid overfitting. This enabled us to assess overfitting vs. generalization in the standard way.

\section{Results and Inferences}

The following section examines the representations learned by the models with special attention to when computing the successor function requires crossing a place-value boundary.

\subsection{Count list model learning}

\begin{figure}[h]
\centering
\includegraphics[width=0.9\columnwidth]{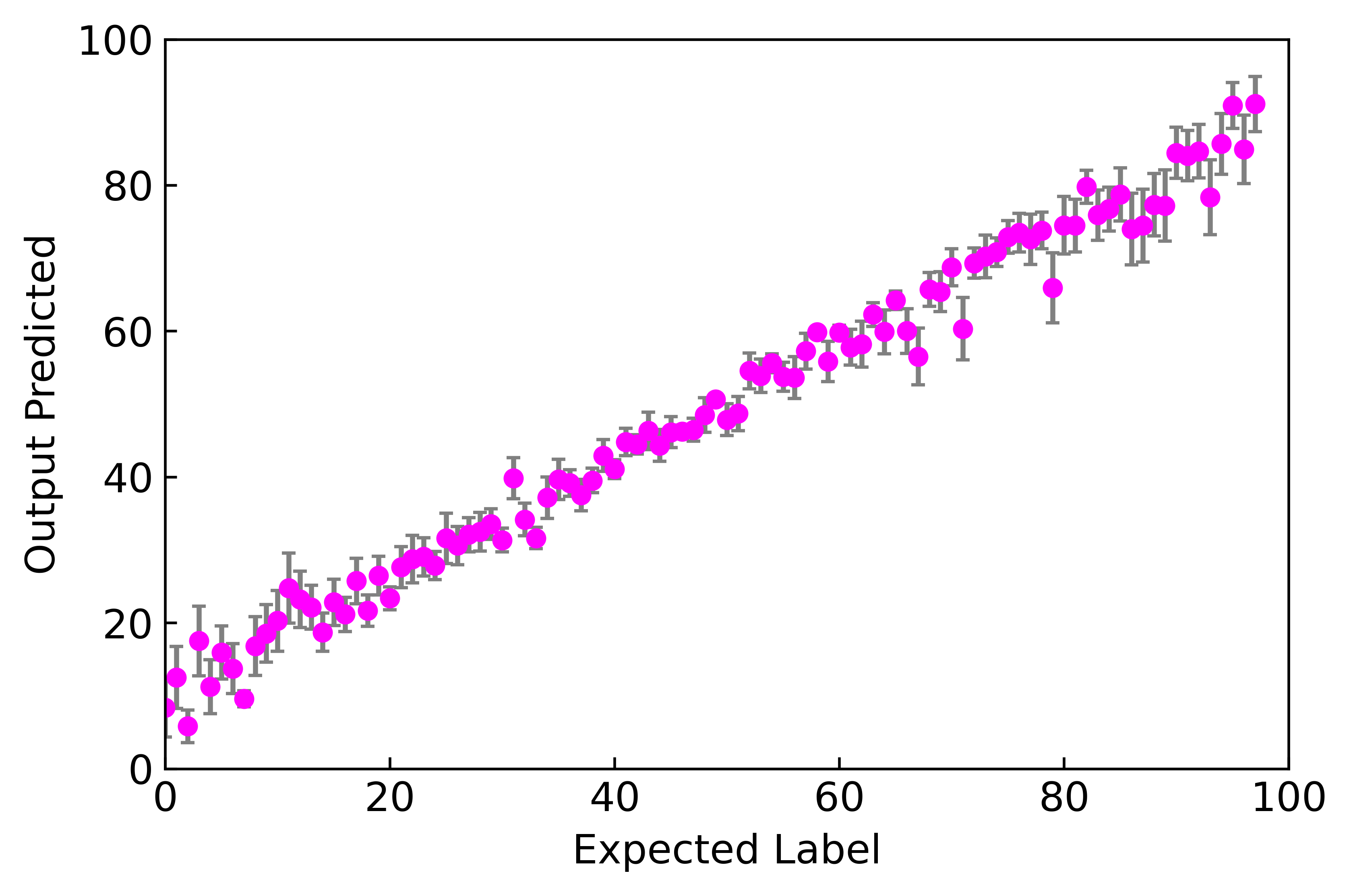}
\caption{Accuracy plot for the count list model.}
\label{fig:acc_one_hot}
\end{figure}

We first evaluated how accurately the count list model computes the successor function using a simple linear regression where the correct successor was the independent variable and the model-predicted successor was the dependent variable. Fig.~\ref{fig:acc_one_hot} shows the output for different values (irrespective of whether they were sampled for training or testing across runs), the model correctly displayed a strong linear relationship between these two variables ($R^2 = .986$). However, its performance was problematic in an absolute sense: the $y$-intercept was far from 0 ($B0 = -11.036$) and the slope was far from 1 ($B1 = 1.24$).

The count list model achieved perfect training accuracy (i.e., 100\% correct computation of the successor) on all of the 25 simulations. However, it showed no generalized understanding of the successor function: test accuracy was 0\% on all 25 simulations as well.

\begin{figure}[h]
\centering
\includegraphics[width=0.9\columnwidth]{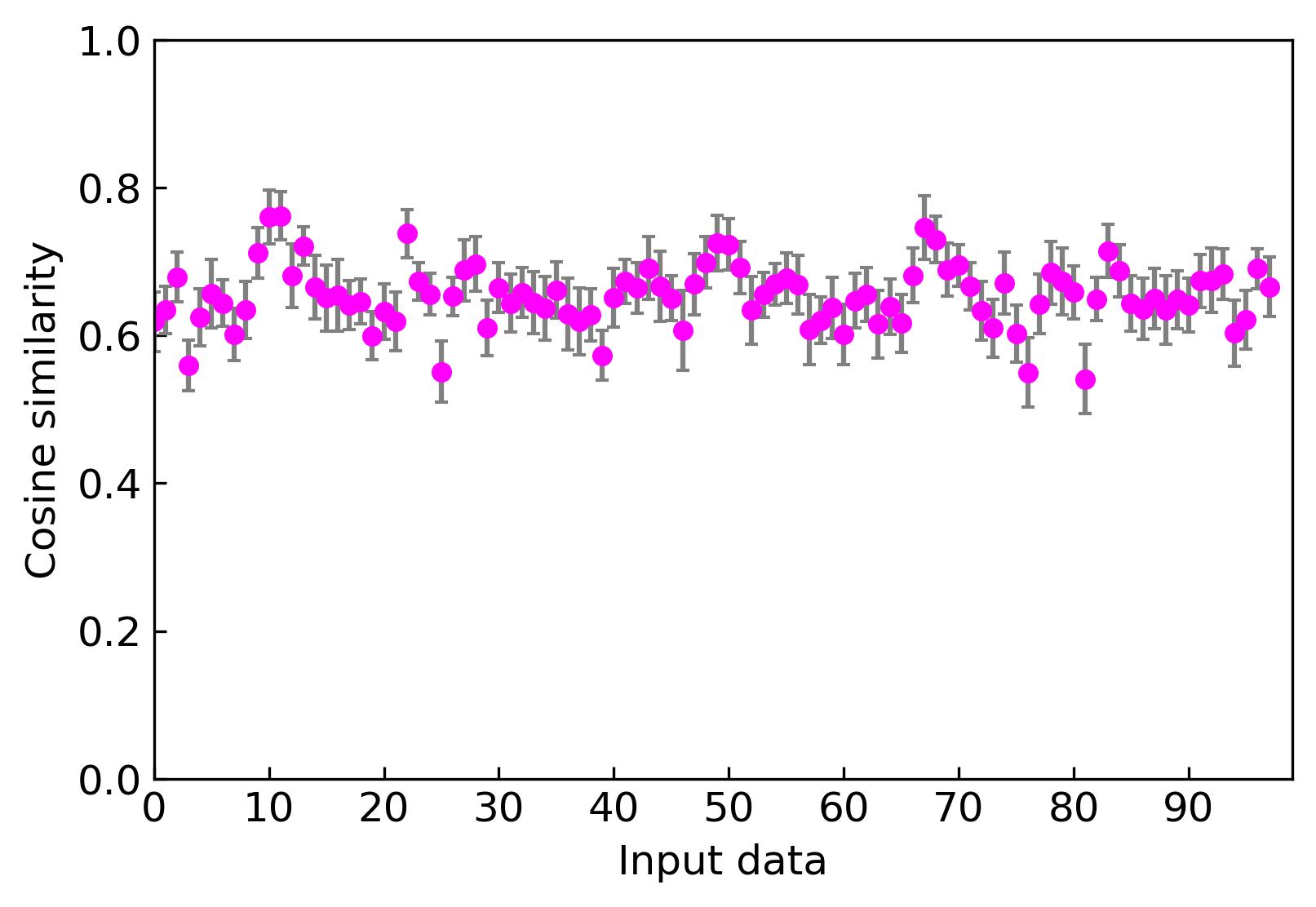}
\caption{Successive cosine similarities for count list model.}
\label{fig:cos_one_hot}
\end{figure}

Given the ordinal structure of natural numbers, a cognitively plausible model should the representation of each number and its successor should be similar. We visualized this for the count list model, plotting the average cosine similarity between each number and its successor; see Fig.~\ref{fig:cos_one_hot}. These similarities were ``medium'' on average (0.654) and their standard deviation was ``small'' (0.024). (We make these terms more precise below when we compare the performance of the count list model to that of the place-value model.) Here, we simply note that the variability in these similarities was not systematic across numbers in a mathematically interesting way.

Given the place-value structure of numbers and its reflection in the numerical syntax of language, a cognitively plausible model should utilize relatively different representations when a number and its successor span a place value (i.e., 10s) boundary ($M = .652, SD = .055$) and relatively similar representations when they do not ($M = .654, SD = .026$). In fact, this was not the case: the average similarity values in these cases were comparable ($t(48) = .15, p = .437$).

\subsection{Place-value model learning}

\begin{figure}[h]
\centering
\includegraphics[width=0.9\columnwidth]{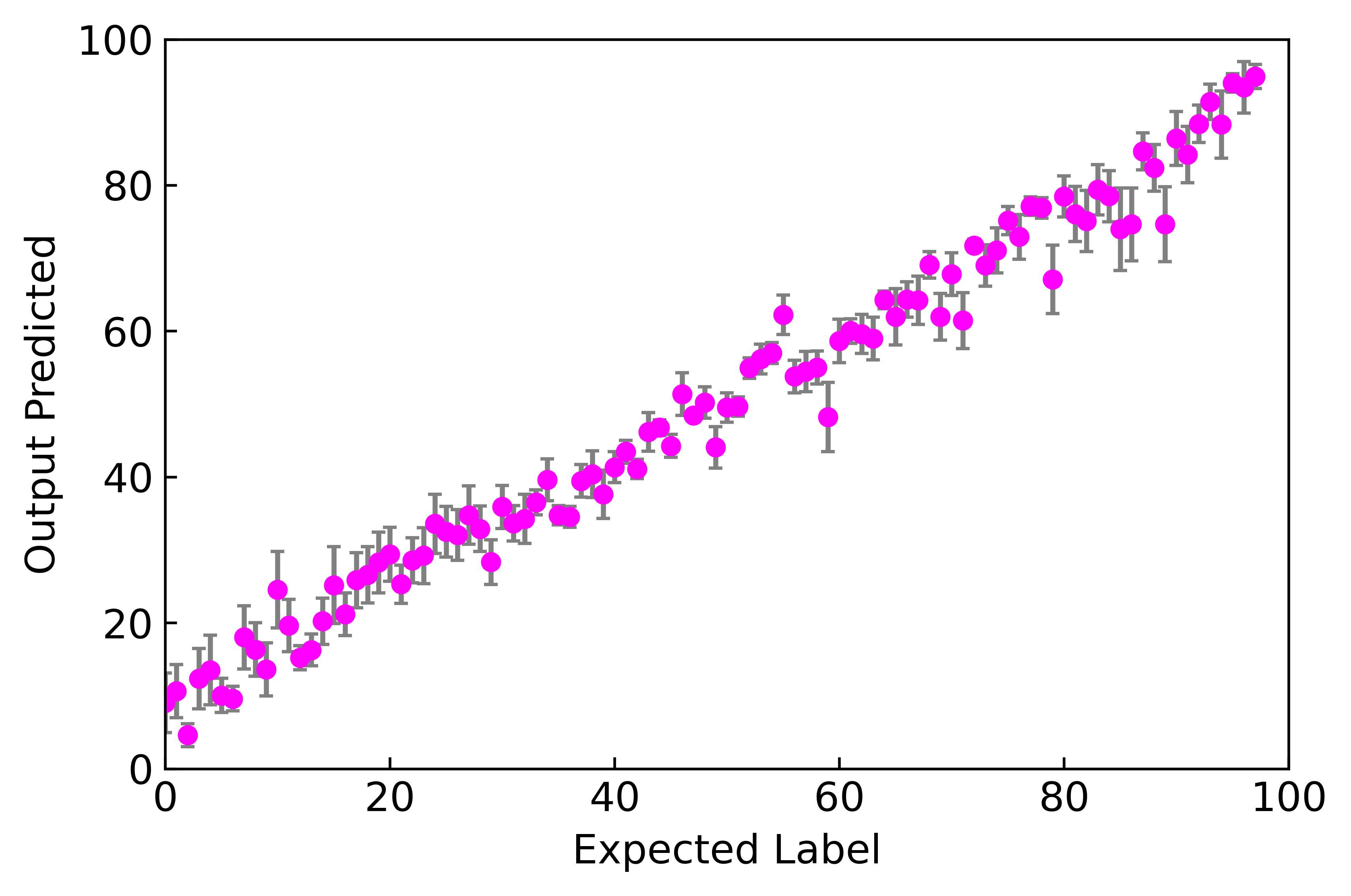}
\caption{Accuracy plot for place value model.}
\label{fig:acc_pv}
\end{figure}

As for the count list model, the linear relationship between the correct successor and the model-predicted successor was very strong for the place-value model ($R^2 = .979$); see Fig.~\ref{fig:acc_pv}. Moreover, the regression coefficients indicated a closer approximation of the true function in an absolute sense: the $y$-intercept was closer to 0 ($B0 = -8.66$ vs. $-11.03$ for the count list model) and the slope closer to 1 ($B1 = 1.17$ vs. $1.24$).

The place-value model was somewhat less accurate than the count list model in terms of training accuracy: 93\% vs. 100\%. Importantly, it showed some generalization of the successor function, with 24\% accuracy on the test data, whereas the count list model showed none (0\%).

\begin{figure}[h]
\centering
\includegraphics[width=0.9\columnwidth]{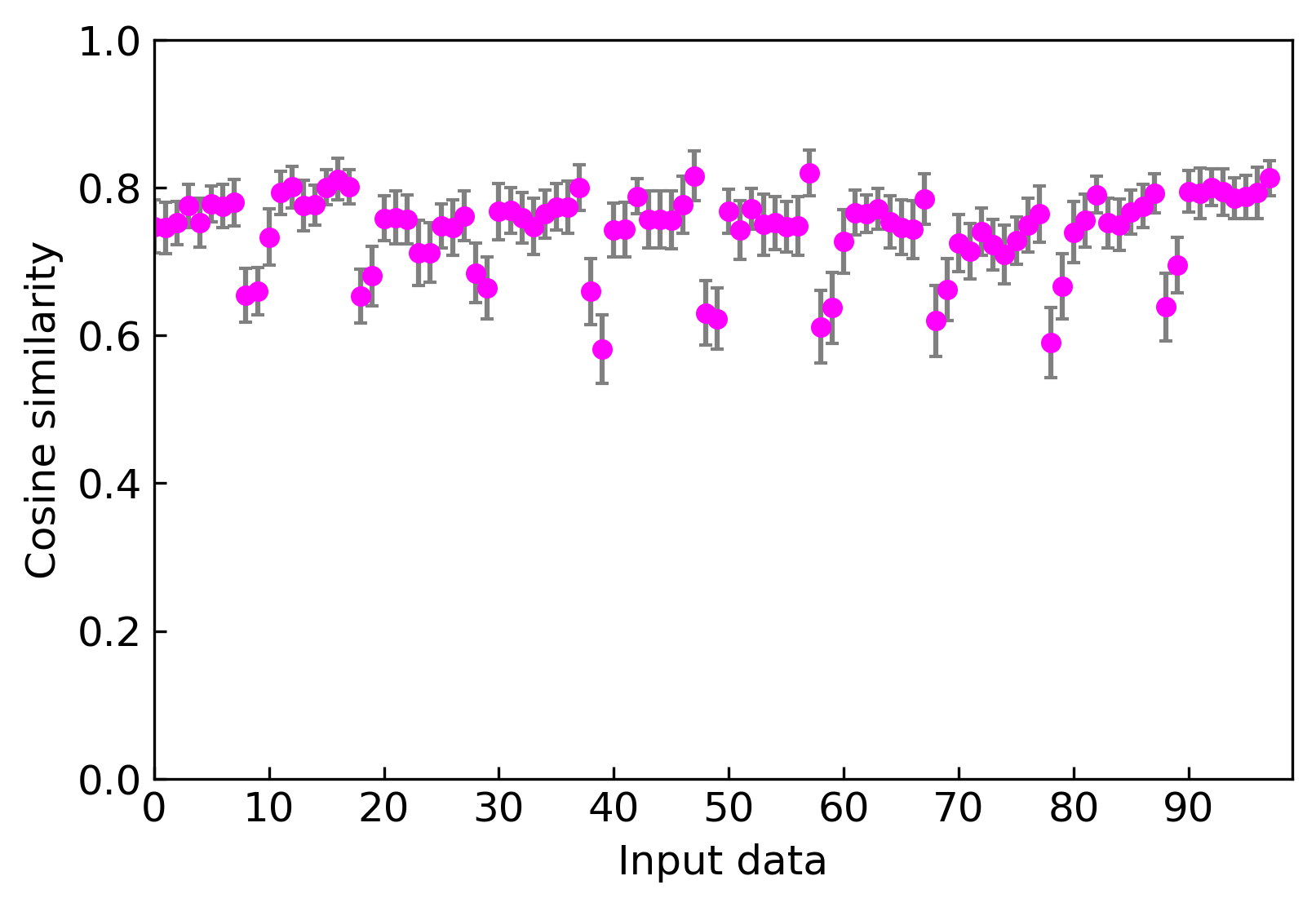}
\caption{Successive cosine similarity plot for the place-value model showing a recurring pattern at the boundaries.}
\label{fig:cos_pv}
\end{figure}

To visualize the place-value model's capturing of the ordinal structure of the natural numbers, we plotted the average cosine similarity between each number and its successor; see Fig.~\ref{fig:cos_pv}. These similarities ($M = .742, SD = .062$) were higher than for the count list model ($t(48) = 6.69, p < .001)$).

We next evaluated whether the average cosine similarity was lower when a number and its successor spanned a place value (i.e., 10s) boundary ($M = .652, SD = .153$) compared to when they were both on the same side ($M = .742, SD = .062$). In fact, this was the case ($t(48) = 2.73, p = .004$). Thus, the place-value model found it useful to internally represent the place value components of numbers when learning to compute the successor function.

\subsection{Latent place-value structure learned by the models}

We analyzed what the two models had learned of the place-value boundaries that are critical for correctly naming the successor of a number \citeA{Miller1995PreschoolOO}. We used MDS to reduce the representations of numbers on the final hidden layer (consisting of 8 units in each model) to 2 dimensions and plotted the vector connecting each number *9 to its successor *0. For a given simulation, a systematic mapping would correspond to vectors of roughly the same angle for each pair of numbers 9 to 10, 19 to 20, ..., 89 to 90. Quantitatively, the standard deviation of these angles should be close to 0. A systematic mapping would also correspond to vectors of large magnitude for each pair of numbers 9 to 10, 19 to 20, ..., and 89 to 90, on average, representing a large jump to a qualitatively different part of representational space. We looked for these two properties in the each of the models.

\begin{figure}[h]
\centering
\includegraphics[width=0.9\columnwidth]{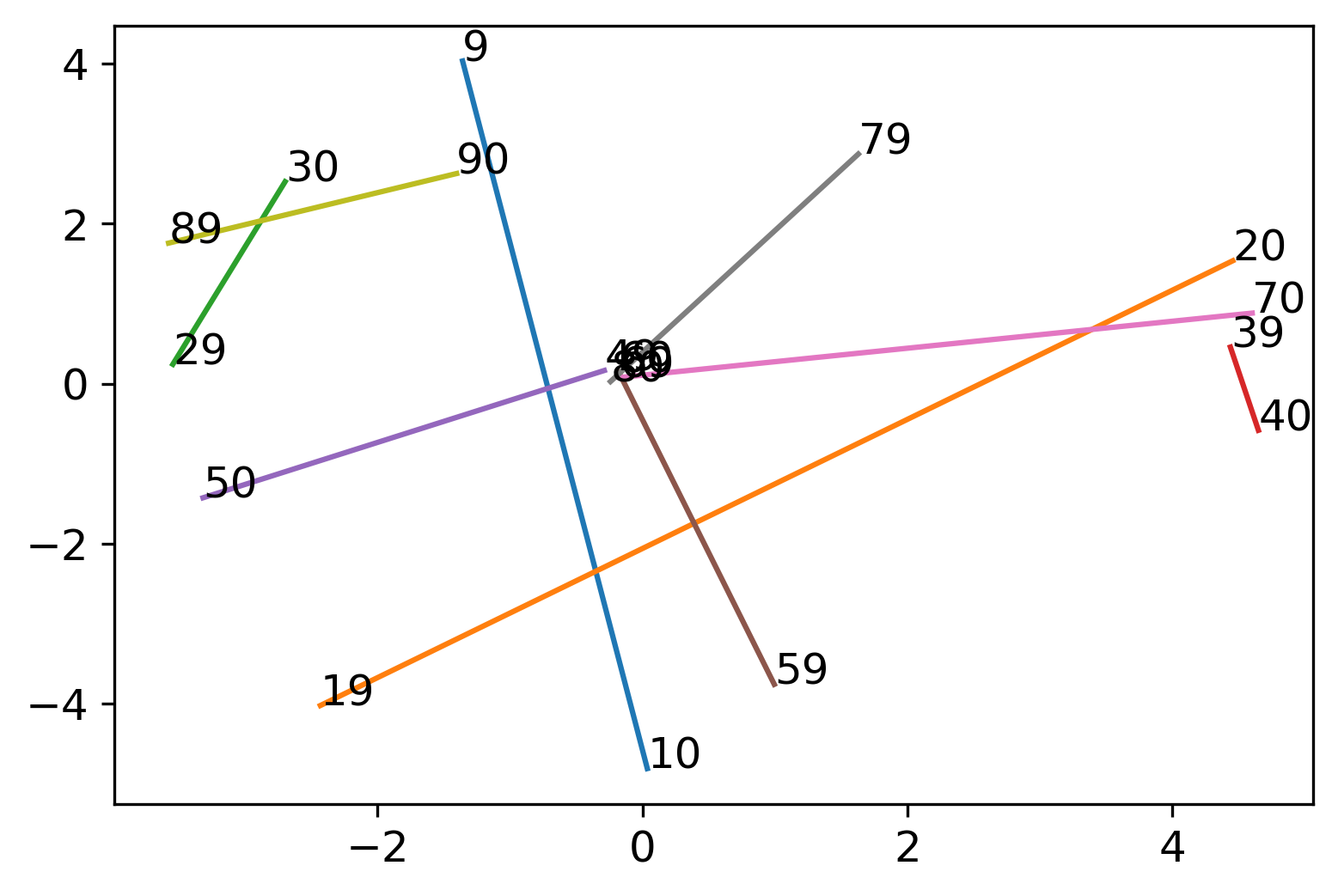}
\caption{Plots of the hidden layer representations of the numbers *9 and *0, reduced by MDS to 2 dimensions, for the count list model, showcasing overlapping representations}
\label{fig:mds_one_hot}
\end{figure}

\begin{figure}[h]
\centering
\includegraphics[width=1\columnwidth]{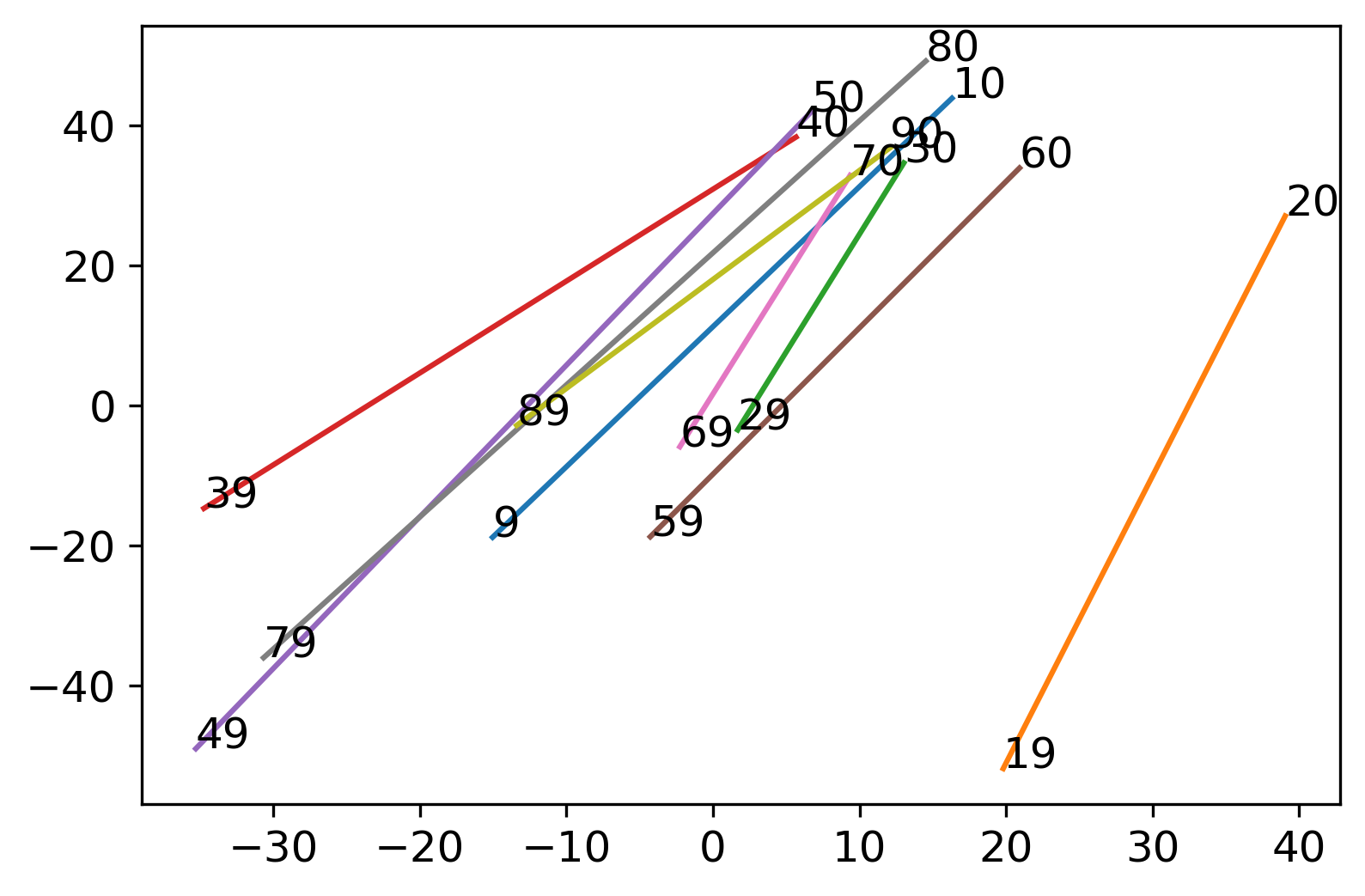}
\caption{Plots of the third hidden layer's representations of the numbers *9 and *0, reduced by MDS to 2 dimensions, for the place-value model, showcasing a systematic mapping of numbers to their successors across the 10s boundaries.}
\label{fig:mds_pv}
\end{figure}

First, consider the hidden layer representations for one of the count list model simulations shown in Fig.~\ref{fig:mds_one_hot}. The angles of the vectors vary greatly and their magnitudes are relatively small, evidencing a lack of systematicity in the mapping of numbers to their successors across 10s boundaries. Fig.~\ref{fig:mds_pv} shows the representations of the third hidden layer of the one of the place-value model simulations. By contrast, the angles of the vectors are approximately the same and the magnitudes of the vectors are relatively large, indicating a systematic mapping of numbers to their successors across 10s boundaries.

\begin{figure}[h]
\centering
\includegraphics[width=0.9\columnwidth]{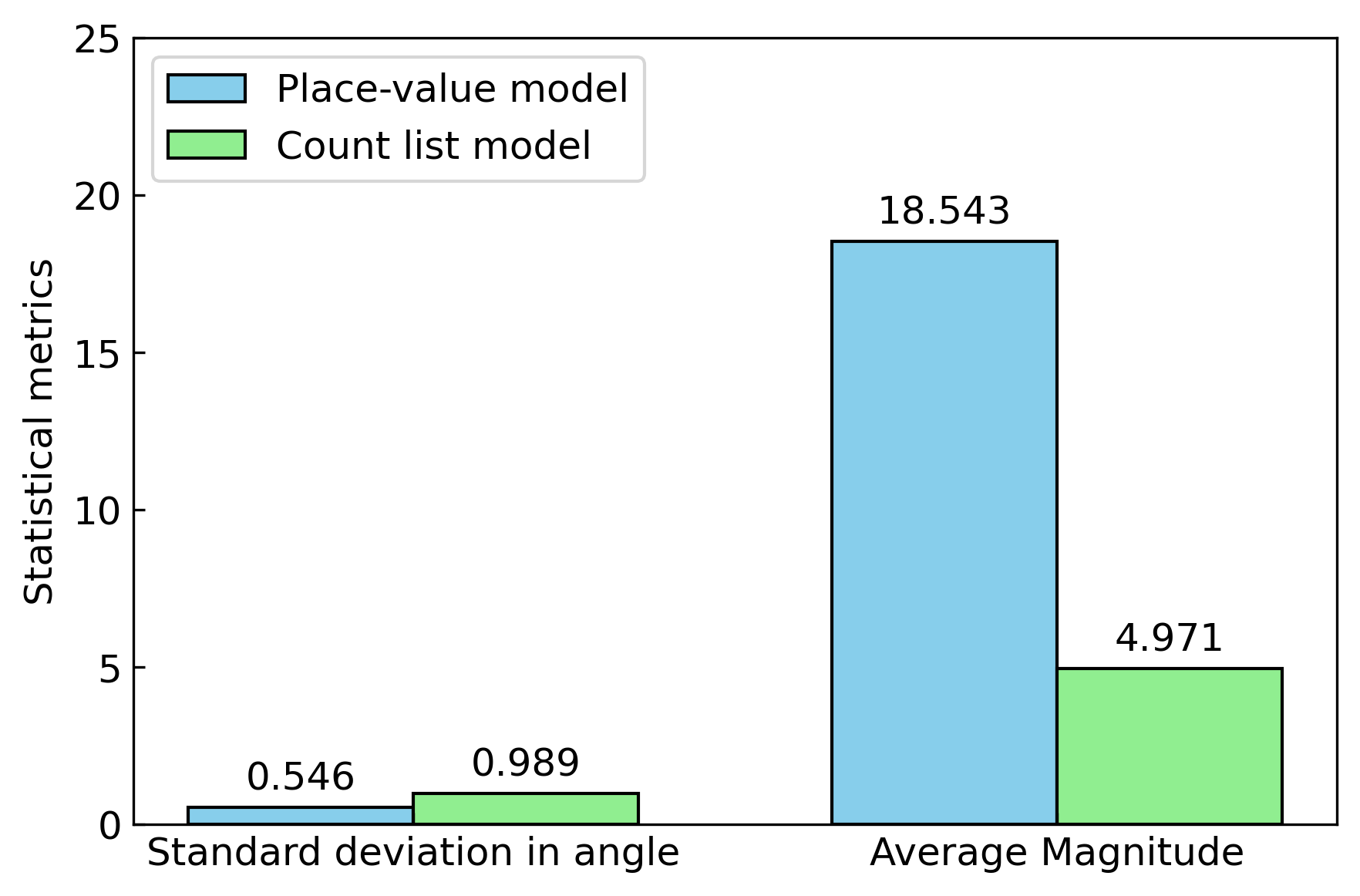}
\caption{The standard deviation of the angles of the vectors formed across the 10s boundaries is lower for the place-value model while the average magnitude of vectors is higher for the place-value model.}
\label{fig:mds_angle}
\end{figure}

An analysis of the 25 simulations of each model verified these observations. We first computed the average standard deviation of the vector angles across the boundary-crossing pairs for the count list model simulations ($M = .989, SD = .118$) radians and the place-value model simulations ($M = .546, SD = .453$) radians. The average standard deviation was smaller for the place-value model ($t(48) = 4.73, p < .001$), indicating a more systematic mapping; see Fig.~\ref{fig:mds_angle}.

We repeated this analysis for the average magnitudes of these vectors for the 25 count list ($M = 4.971, SD = .880$) and the 25 place-value ($M = 18.513, SD = 9.310$) model simulations. As shown in  Fig.~\ref{fig:mds_angle} this magnitude was greater for the place-value model ($t(48) = 7.239, p < .001$), again indicating a more systematic mapping.

\subsection{Curriculum learning of the successor function}
\label{subsec:curr}
To more closely simulate numerical development, we took a curriculum learning approach. We again focused on the place-value model because of its superior performance to the count list model in the analyses above. Rather than providing the entire training data set beginning in the first epoch, we gradually expanded the size of the training data set by adding the next 20 numbers every 1000 epochs. Thus, for the first 1000 epochs, the model was trained on the data set $D:[0,19] \rightarrow R:[1,20]$. For the next 1000 epochs, it was trained on the expanded data set $D:[0,39] \rightarrow R:[1,40]$. This process was repeated until the model was trained for the final 1000 epochs on the full data set $D:[0,98] \rightarrow R:[1,99]$. Finally, we ran a sixth simulation where we trained the model for an additional 1000 epochs on the entire dataset to be fair to the newly acquired data. As in prior modeling, we repeated the curriculum learning simulations 25 times and considered the average performance across these simulations.

\begin{figure}[h!]
\centering
\includegraphics[width=1\columnwidth]{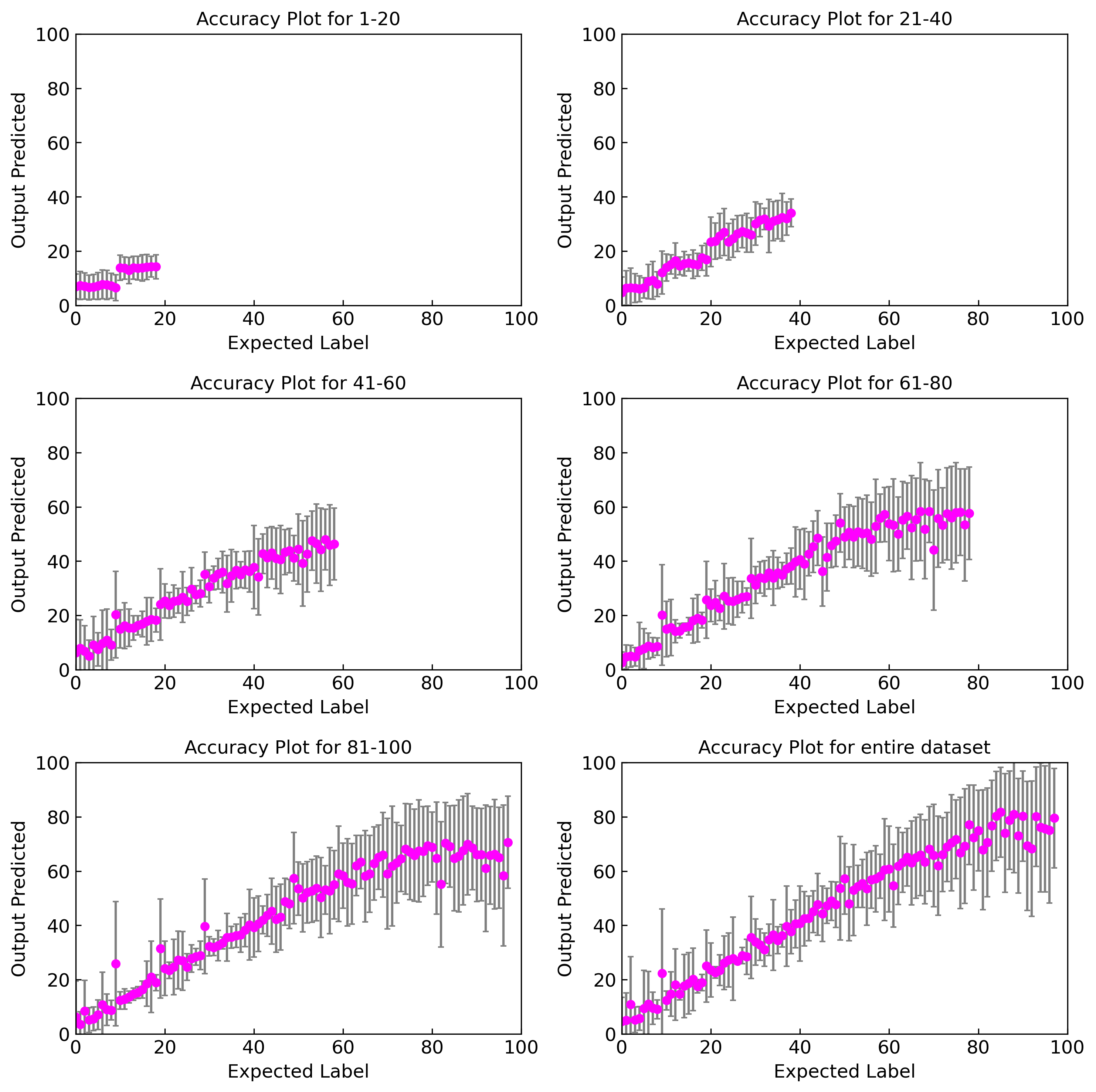}
\caption{Curriculum learning simulation performance plots.}
\label{fig:curr}
\end{figure}

Figure~\ref{fig:curr} shows the performance of the model every 1000 epochs for the training data to that point. As the data set grows, the model incrementally learns the successor function. Importantly, it is increasingly accurate (i.e., more precise, less variable) for smaller numbers, which it has trained on for many epochs. This is especially the case for the single-digit numbers, which it has trained on for the most epochs, but which also come to represent a smaller and smaller proportion of the training data. By contrast, the model's accuracy for larger number increases more slowly.

\begin{figure}[h!]
\centering
\includegraphics[width=0.9\columnwidth]{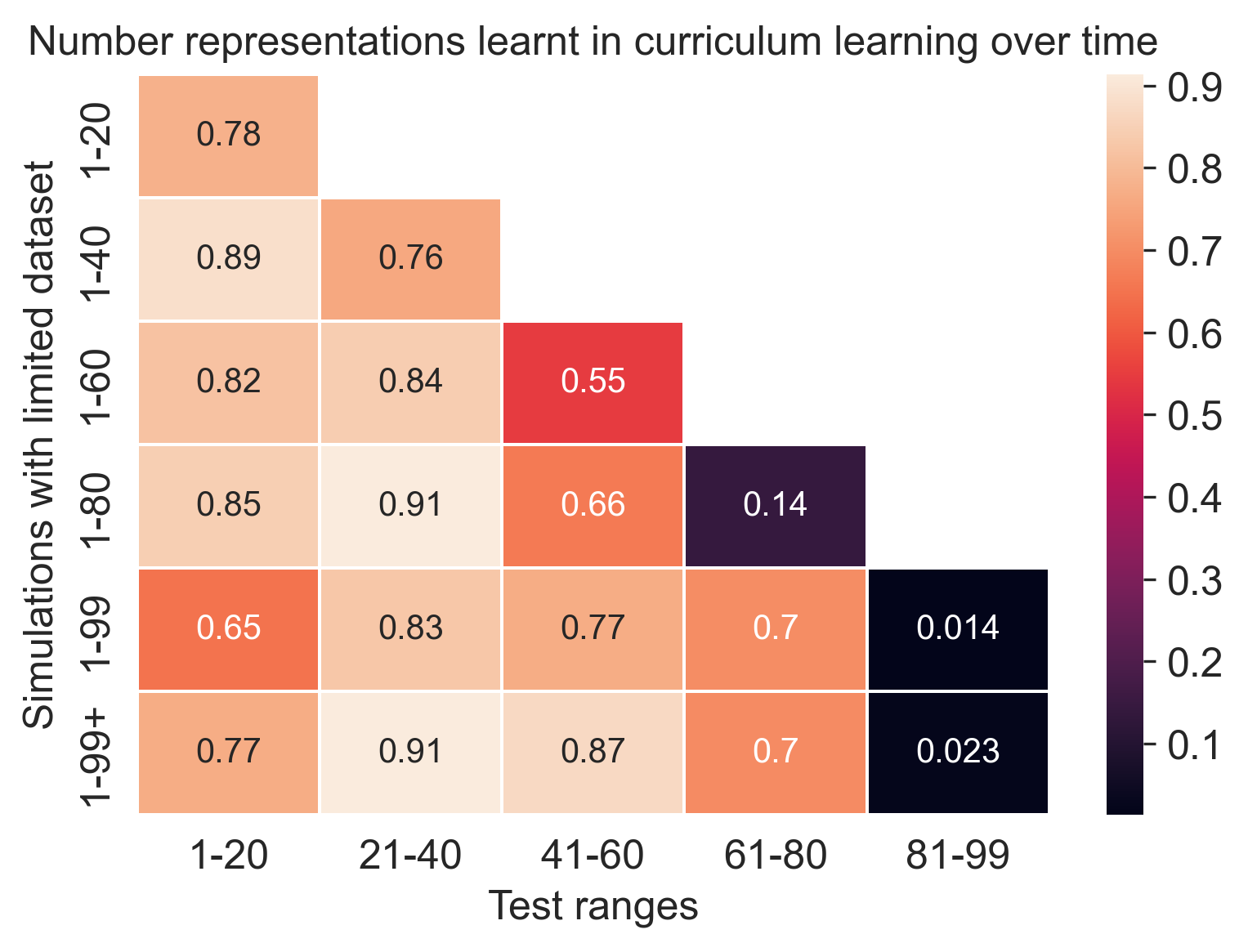}
\caption{Curriculum learning analysis.}
\label{fig:curr_map}
\end{figure}

Finally, we evaluated the sharpness of the number representations learned over time in the curriculum learning simulations. For each set of 25 simulations for which a range was relevant, we computed the linear fit separately for numbers in the test ranges 1 – 20, 21 – 40, 41 – 60, 61 – 80, and 81 – 99. Specifically, we computed the correlation between the correct successor and the average model-predicted successor. These correlations are shown in Fig.~\ref{fig:curr_map}. The rows correspond to simulations over expanding subsets of the training data. The columns organize the different test ranges. There are three interesting patterns to notice. First, the correlation generally increases for a test range the longer the model trains on it. Second, as the size of the training data subset increases, we observe a saturation in the correlation, and occasional dips as well (and there is pronounced non-linearity for the test range 1 – 20). Third, the correlations between the correct successor and the average model-predicted successor are quite low for the highest test range 81-99. This is true even for the final set of simulations where we trained the model for an additional 1000 epochs to enable better learning of this test range.

\section{Discussion}
This study investigated the number representations of two neural network models trained on the successor function. This is the first step to taking a neural networks approach to an important question in cognitive science \citeA{carey2019ontogenetic, doi:10.1080/09515080802285354, SPELKE2017} and the philosophy of mathematics more generally \citeA{giaquinto2001knowing}: What does it mean to learn the successor function, mechanistically, and ultimately to understand the countably infinite?

The count list model learned a continuous and approximate notion of number. It was not accurate on the testing data ($0\%$) and showed no sensitivity to tens boundaries. This is an important limitation because learning these boundaries is critical for the place-value structure and numerical syntax of naming the successor of a number \citeA{Guerrero2020IsTT, Miller1995PreschoolOO, Pollmann1996TheLU, Schneider2020DoCU}.

The place-value model was more successful. Its accuracy on the testing data was higher ($24\%$) -- though far from perfect -- and its number representations were sensitive to tens boundaries. An MDS analysis of the hidden layer representations revealed that generating the successor across a tens boundary could be visualized as vector addition in 2D space.
These experiments probing the latent representations learned by the place-value model offer mechanistic insights into what it means for children to understand the successor function. That said, the models learn formally different ML tasks, making it difficult to directly compare them.

The curriculum learning simulation explored numerical development in the expanding numerical environment of the developing child \citeA{Miller1995PreschoolOO}. It revealed sharpening representations of smaller numbers even as larger numbers became part of the training data. These results suggest that learning single-digit numbers may bootstrap learning of multi-digit numbers, and of place-value more generally. However, the models did not learn the successor function for the highest test range 81 – 99, not even the final set of simulations where the model was trained for an additional 1000 epochs on the full dataset. That this was not successful suggests that learning a given test range may well require also learning higher test ranges or changing model architecture to promote generalization~\citeA{Nam2022OutofDistributionGI}.

\section{Conclusion}
There are important differences between how neural networks learn and how children develop. Most neural networks learn within the supervised paradigm, whereas children make substantial use of unsupervised learning  (called ``statistical'' learning in the developmental literature).
In addition, neural networks typically require orders of magnitude more training samples than children. Future research should move beyond generating the successor of a number to modeling the counting process more generally. We are exploring training recurrent neural network models to sequentially count, first to 100 and then arbitrarily high for out-of-distribution data. Prior ML and cognitive science research has applied recursive neural network architectures to learn to count numbers \citeA{861302} and sets \citeA{fang2018can}. Our hypothesis is that (1) training on the more complex counting task and (2) using richer recurrent architectures will enable models to learn the recursive structure of the natural numbers, yielding more promising candidates for what it is that children learn when they come to understand the countably infinite.

\bibliographystyle{apacite}

\setlength{\bibleftmargin}{.125in}
\setlength{\bibindent}{-\bibleftmargin}

\bibliography{CogSci_Template}

\end{document}